\documentclass[runningheads]{llncs}
\usepackage{graphicx}
\usepackage{amsmath}
\usepackage{amssymb}
\usepackage{tabulary}
\usepackage{booktabs}
\usepackage{cite}
\usepackage[numbers,sort&compress]{natbib}
\usepackage[hidelinks]{hyperref}
\usepackage{lscape}
\usepackage{rotating}
\usepackage{afterpage}

\begin{document}
\title{Explaining Chest X-ray Pathologies in \\ Natural Language}

%
%
\author{Maxime Kayser\inst{1}\thanks{Corresponding author: firstname.lastname@cs.ox.ac.uk} \and 
Cornelius Emde\inst{1} \and 
Oana-Maria Camburu\inst{2}\thanks{This work was mostly done while Oana was at the University of Oxford.} \and 
Guy Parsons\inst{1,3} \and 
Bartlomiej Papiez\inst{1} \and 
Thomas Lukasiewicz\inst{4,1}} 

%
\authorrunning{M. Kayser, C. Emde, O. Camburu, G. Parsons, B. Papiez, T. Lukasiewicz}
%
\institute{University of Oxford, United Kingdom\\
\and 
University College London, United Kingdom \and
Thames Valley Deanery, Oxford, United Kingdom \and 
TU Wien, Austria}
\maketitle              
\begin{abstract}
Most deep learning algorithms lack explanations for their predictions, 
which limits their deployment in clinical practice. Approaches to improve explainability, especially in medical imaging, 
have often been shown to convey limited information, be overly reassuring, or lack robustness. 
In this work, we introduce the task of generating natural language explanations (NLEs) to justify predictions made on medical images. 
NLEs are human-friendly and  comprehensive, 
and enable the training of intrinsically explainable models. To this goal, we introduce MIMIC-NLE, the first, large-scale, medical imaging dataset with NLEs. It contains over 38,000 NLEs, which explain the presence of various thoracic pathologies and chest X-ray findings. We propose a general approach to solve the task and evaluate several architectures on this dataset, including via clinician assessment. 

\keywords{Chest X-rays  \and Natural Language Explanations \and XAI}
\end{abstract}
\section{Introduction} \label{sec:intro}
Deep learning (DL) has become the bedrock of modern computer vision algorithms. However, a major hurdle to adoption and regulatory approval of DL models in medical imaging is the lack of explanations for these models' predictions~\cite{langlotz_roadmap_2019}. The combination of lack of model robustness \cite{papernot_transferability_2016}, bias (algorithms are prone to amplifying inequalities that exist in the world) \cite{obermeyer_dissecting_2019, hajian_algorithmic_2016}, and the high stakes in clinical applications \cite{vayena_machine_2018, mozaffari-kermani_systematic_2015} prevent black-box DL algorithms from being used in practice. 
In this work, we propose natural language explanations (NLEs) as a means to justify the predictions 
of medical imaging algorithms.



So far, the most commonly used form of explainability in medical imaging is saliency maps, which attribute importance weights to regions in an image. Saliency maps have many shortcomings, including being susceptible to adversarial attacks \cite{gu_saliency_2019}, conveying limited information and being prone to confirmation bias \cite{adebayo_sanity_2018, bornstein_is_2016}, as well as only telling us \emph{how much} highlighted regions affect the model's output, and not \emph{why} \cite{ghassemi_false_2021}. NLEs, on the other hand, would be able to fully capture how the evidence in a scan relates to the diagnosis. Furthermore, saliency maps are post-hoc explainers, i.e., they do not constrain the model to learn in an explainable manner. In contrast, self-explaining models have many benefits, including being more robust and having a better prediction performance \cite{Stacey22}. %
Alternative 
approaches for explainability in medical imaging include latent space disentanglement \cite{puyol-anton_interpretable_2020}, counterfactual explanations \cite{schutte_using_2021}, case-based explanations \cite{kim_xprotonet_2021}, and concept-based explanations \cite{koh_concept_2020}. NLEs are a valuable addition to the suite of self-explaining models for medical imaging, as they provide easy-to-understand explanations that are able to communicate complex decision-making processes and mimic the way in which radiologists explain diagnoses~\cite{miller_explanation_2019, gale_producing_2018}.
Previous attempts to augment medical image classification with textual information rely on template-generated sentences \cite{gale_producing_2018, lee_generation_2019} or sentences from other images based on image similarity \cite{kougia_rtex_2021}. Furthermore, their focus lies mostly on adding descriptive information about a pathology (e.g., its location), instead of explaining the diagnoses. In this work, we leverage the free-text nature of chest X-ray radiology reports, which 
do not only provide additional details about pathologies, but also the degree of certainty of a diagnosis, as well as justifications of how other observations explain it. Our work builds on the growing work on NLEs approaches in natural language processing \cite{camburu_e-snli_2018, kotonya_explainable_2020, rajani_explain_2019}, natural image understanding \cite{park_multimodal_2018, marasovic_natural_2020, kayser_e-vil_2021, majumder_rationale-inspired_2021}, as well as task-oriented tasks such as self-driving cars \cite{kim_textual_2018} and fact-checking \citep{kotonya_explainable_2020}.

We propose MIMIC-NLE, the first dataset of NLEs in the medical domain. MIMIC-NLE extends the existing MIMIC-CXR dataset of chest X-rays \cite{johnson_mimic-cxr-jpg_2019} with diagnoses, evidence labels and NLEs for the diagnoses. 
We create MIMIC-NLE by using a BERT-based labeler, a set of clinical explanation keywords, and an empirically and clinically validated set of extraction rules. We extracted over 38,000 high-quality NLEs from the over 200,000 radiology reports present in MIMIC-CXR. Our extraction process introduces little noise, on-par or better than for NLE datasets in natural images \cite{kayser_e-vil_2021}. 
Second, we establish an evaluation framework and compare three strong baselines. 
The evaluation by a clinician validates the feasibility of the task, but also shows that it is a challenging task requiring future research. The code and dataset are publicly available at \url{https://github.com/maximek3/MIMIC-NLE}.
\section{MIMIC-NLE}
\begin{figure}
\begin{center}
\includegraphics[width=1\linewidth]{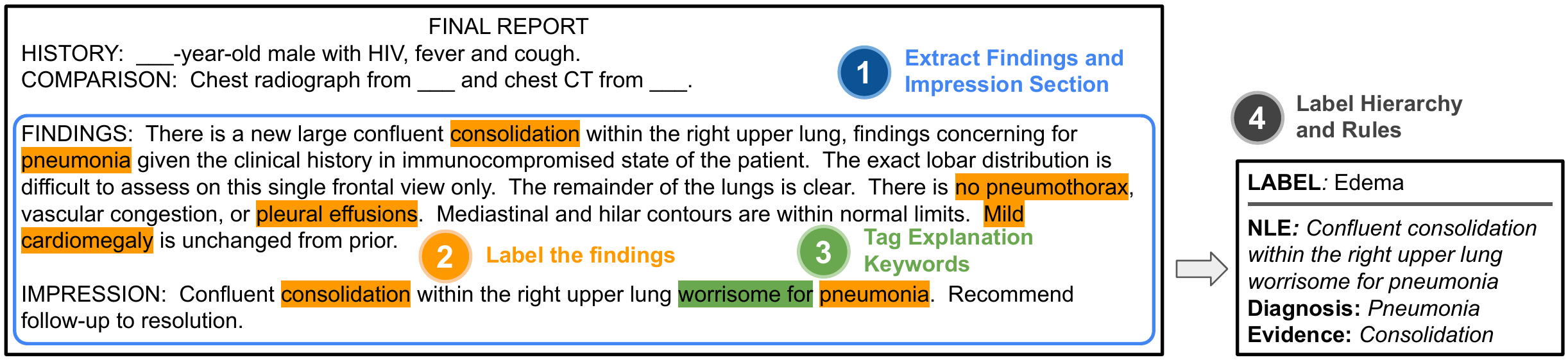}
\end{center}
    \caption{The steps required to extract NLEs from raw radiology reports. We first extract the Findings and Impression sections, which contain the descriptive part of the report. Next, we identify the labels referred to in each sentence and the sentences that contain explanation keywords. Based on this information, we leverage the rules described in Table~\ref{tab:lbl_rules} to extract valid NLEs, as well as their diagnosis and evidence labels.}%
    \label{fig:nle_extraction}
\end{figure}

Gathering NLEs is expensive, especially when radiologic expertise is required. To our knowledge, there is currently no NLE dataset for medical imaging. To address this, we show that it is possible to automatically distill NLEs from radiology reports, as radiologists typically explain their findings in these reports. We leverage the radiology reports from MIMIC-CXR \cite{johnson_mimic-cxr-jpg_2019}, a publicly available chest X-ray dataset with 227,827 radiology reports. By applying various filters, labelers, and label hierarchies, we extract 38,003 \emph{image-NLE} pairs, or 44,935 \emph{image-diagnosis-NLE} triplets (as some NLEs explain multiple diagnoses). The extraction process is summarized in Fig.~\ref{fig:nle_extraction}. Our filters consist mainly of removing sentences that contain anonymized data, or provide explanations based on patient history or technical details of the scan. More details are provided in the appendix. Furthermore, we only consider frontal, i.e., anteroposterior (AP) and posteroanterior (PA) scans, as these are most commonly used for diagnosis in routine clinical pathways and are most likely to contain the visual information required to generate NLEs.

\begin{figure}
\begin{center}
\includegraphics[width=1\linewidth]{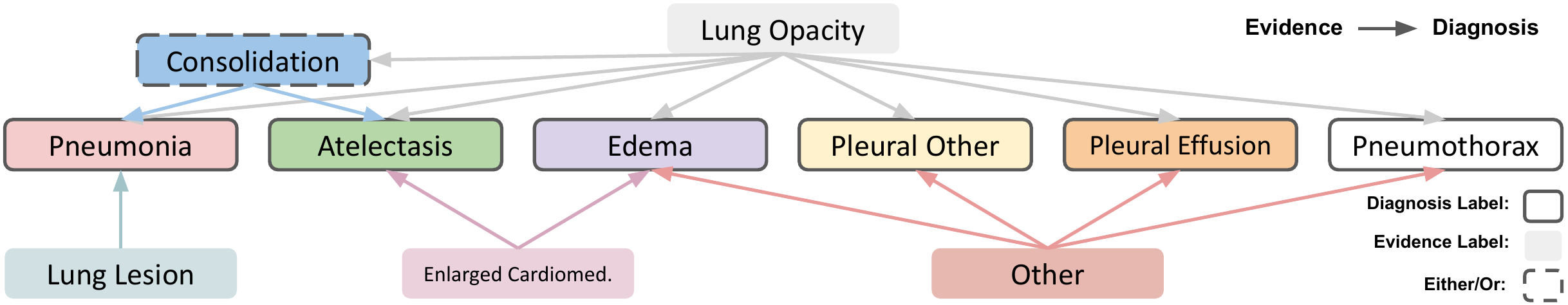}
\end{center}
    \caption{Our evidence graph that visualizes which of the labels can act as evidence for which diagnosis labels.}%
    \label{fig:evidence_graph}
\end{figure}

First, based on exploring the reports and discussions with clinicians, we observe that a small selection of phrases, such as ``compatible with'' or ``worrisome for'' (see full list in Appendix \ref{app:data}), are very accurate identifiers of sentences where a potential pathology is explained by observations made on the scan (i.e., an \emph{explanatory} sentence). Next, we make use of the CheXbert labeler \cite{smit_combining_2020}, which can extract 14 different chest X-ray labels from clinical text, to identify the findings mentioned in each sentence. Thus, for each sentence in the MIMIC-CXR radiology reports, we know whether it is an explanatory sentence, which labels it refers to, and if the labels have a negative, positive, or uncertain (i.e., they are \emph{maybe} present) mention. As the goal of NLEs is to explain predictions, we need to establish which of the labels mentioned in an NLE are being explained and which are part of the evidence. To be able to determine the evidence relationships present in NLEs, we need to restrict ourselves to a limited set of label combinations. For example, if an explanatory sentence in a radiology report indicates the presence of both \emph{Atelectasis} and \emph{Consolidation}, it is not obvious whether \emph{Consolidation} is evidence for \emph{Atelectasis} (e.g., ``Right upper lobe new consolidation is compatible with atelectasis with possibly superimposed aspiration.'') or whether they are both explained by a different finding (e.g., ``A persistent left retrocardiac density is again seen reflecting left lower lobe atelectasis or consolidation.''). However, for other label combinations, such as \emph{Consolidation} and \emph{Pneumonia}, the evidence relationship is usually clear, i.e., \emph{Consolidation} is the evidence for \emph{Pneumonia}. We therefore propose the evidence graph in Fig.~\ref{fig:evidence_graph}, which depicts the label relationships that have a known and high-confidence evidence relationship (i.e., their co-occurrence in a sentence lets us deduct an evidence relationship with high probability). The graph was constructed with an external radiologist and by empirically validating the co-occurrences of these labels in MIMIC-CXR. 

Based on the evidence graph and manual inspection of every label combination that appears at least 25 times in an explanatory sentence (121 combinations in total), we established 12 mutually exclusive rules that mark a sentence as a valid NLE. The rules take into account the label combination, their uncertainty label (i.e., \emph{Uncertain} or \emph{Positive}), and the presence of explanation keywords. The rules are defined in Table~\ref{tab:lbl_rules}. For each sentence in MIMIC-CXR, we use these rules to determine whether the label combination referred to in the sentence makes it a valid NLE and to determine which of the labels are part of the evidence and the diagnosis. Some label combinations, such as \emph{Consolidation} and \emph{Pneumonia}, are considered a valid NLE even if they do not contain an explanation keyword (as the keywords are not exhaustive). However, for label combinations with less frequent evidence relationships, such as \emph{Enlarged Cardiomediastinum} and \emph{Edema}, we also require the presence of an explanation keyword. It is worth noting that we focus on explanations for positive and uncertain cases only, as strictly negative findings will generally not require case-specific explanations.

Based on these rules, we obtain 38,003 NLEs. One of the authors evaluated a subset of 100 NLEs and found an accuracy of 92\% (compared to between 49\% to 91\% for natural images NLE datasets \cite{kayser_e-vil_2021}). We deemed an NLE as correct if it correctly uses the extracted evidence labels to explain the extracted diagnosis labels. The main reasons for incorrect NLEs were failures of the CheXbert labeler. Thus, with improved labelers, the accuracy could be further improved. Using the train, dev, and test splits in MIMIC-CXR \cite{johnson_mimic-cxr-jpg_2019}, we get split sizes of 37,016,  273, and 714 for MIMIC-NLE, respectively. 


\section{Models}
We propose self-explaining models that learn to detect lung conditions and explain their reasoning in natural language. The learned image representations are constrained by mapping them to language that explains the evidence backing a diagnosis. Our approach is illustrated in Fig.~\ref{fig:nle_extraction}.

We denote the vision model, i.e., the image classification model, as task model $M_T$. In our case, $M_T(x)=Y$, where $x$ is a radiolographic scan, and $Y$ is the prediction vector $Y \in \mathbb{R}^{n_{\text{unc}} \times n_{\text{path}}}$, with $n_{\text{unc}}=3$ being the number of certainty levels, and $n_{\text{path}}=10$ being the number of pathologies.
This follows the \emph{U-MultiClass} approach from \citet{irvin_chexpert_2019}, i.e., for each pathology we classify the image as \emph{negative}, \emph{uncertain}, or \emph{positive}. 

\begin{table}[]
    \begin{center}
    \caption{This table denotes all the included label combinations for NLEs, including which of the labels are being explained and which are the evidence. The column ``\emph{kw req.}'' specifies which label combinations additionally require the presence of an explanation keyword to be considered an NLE. ``\textit{Other / misc.}'' refers to evidence that has not been picked up by the CheXbert labeler.
    If not denoted by $^U$ or $^P$, all labels can be either positive or uncertain. $A^U$ and $B^U$ are the sets $A$ and $B$, where all labels are given as uncertain. $\mathcal{P}_{\geq2} (A^U)$ is the power set of $A^U$, where each set has at least two labels (i.e., any combination of at least two labels from $A^U$).} \label{tab:lbl_rules}
    \label{tab:rules}
    \begin{tabular}{@{}|l|l|c|@{}}
    \hline
    \multicolumn{2}{|c|}{\textbf{MIMIC-NLE Label Combinations}}    
    &                  \\
    \hline
    \textbf{Evidence}                         & \textbf{Diagnosis Label(s)}                                                                              & \textbf{kw req.} \\
    \hline
    \textit{Other / misc.}                    & $d \in A = \{\mathrm{Pleural \ Eff.}, \mathrm{Edema}, \mathrm{Pleural \ Other}, \mathrm{Pneumoth.}\}$    & yes              \\
    \textit{Other / misc.}                    & $s \in \mathcal{P}_{\geq2} (A^U)$                                                                        & yes              \\
    $\mathrm{Lung \ Opacity}$                 & $d \in B = A \cup \{\mathrm{Pneumonia}, \mathrm{Atelectasis}\}$                                          & no               \\
    $\mathrm{Lung \ Opacity}$                 & $s \in \mathcal{P}_{\geq2} (B^U)$                                                                        & no               \\
    $\mathrm{Lung \ Opacity}$                 & $\mathrm{Consolidation}$                                                                                 & no               \\
    $\mathrm{Consolidation}$                  & $\mathrm{Pneumonia}$                                                                                     & no               \\
    $\{\mathrm{Lung \ Op.}, \mathrm{Cons.}\}$ & $\mathrm{Pneumonia}$                                                                                     & no               \\
    $\mathrm{Lung \ Lesion}$                  & $\mathrm{Pneumonia}$                                                                                     & yes              \\
    $\mathrm{Lung \ Opacity}$                 & $\{\mathrm{Atelectasis^P}, \mathrm{Pneumonia^U}\}$                                                       & no               \\
    $\mathrm{Consolidation}$                  & $\{\mathrm{Atelectasis^U}, \mathrm{Pneumonia^U}\}$                                                       & no               \\
    $\mathrm{Enlarged \ Card.}$               & $\mathrm{Edema}$                                                                                         & yes              \\
    $\mathrm{Enlarged \ Card.}$               & $\mathrm{Atelectasis}$                                                                                   & yes              \\
    \hline
    \end{tabular}
    \end{center}
\end{table}

To our knowledge, this is the first application of NLEs to multi-label classification. We address this by generating an NLE for every label that was predicted as \emph{uncertain} or \emph{positive} and is considered a diagnosis label in our evidence graph. Given a set of pathologies $P$ (see Fig.~\ref{fig:evidence_graph}), we generate an NLE for every pathology $p_j \in P$. We denote the explanation generator as $M_E$. 
For every pathology $p_j$, we condition the NLE generation on the $M_T$ prediction, i.e., we have $M_E(x_{\text{REP}}, Y, p_j) = e_j$, where $x_{\text{REP}}$ is the learned representation of the image $x$, and $e_j$ is the NLE that explains the classification of $p_j$.
By backpropagating the loss of $M_E$ through $M_T$, $M_T$ is constrained to learn representations that embed the correct reasoning for each diagnosis. 
During training, 
we condition $M_E$ on the ground-truth (GT) pathology $p_j$ and prediction vector $Y$.

\begin{figure}
\begin{center}
\includegraphics[scale=0.5]{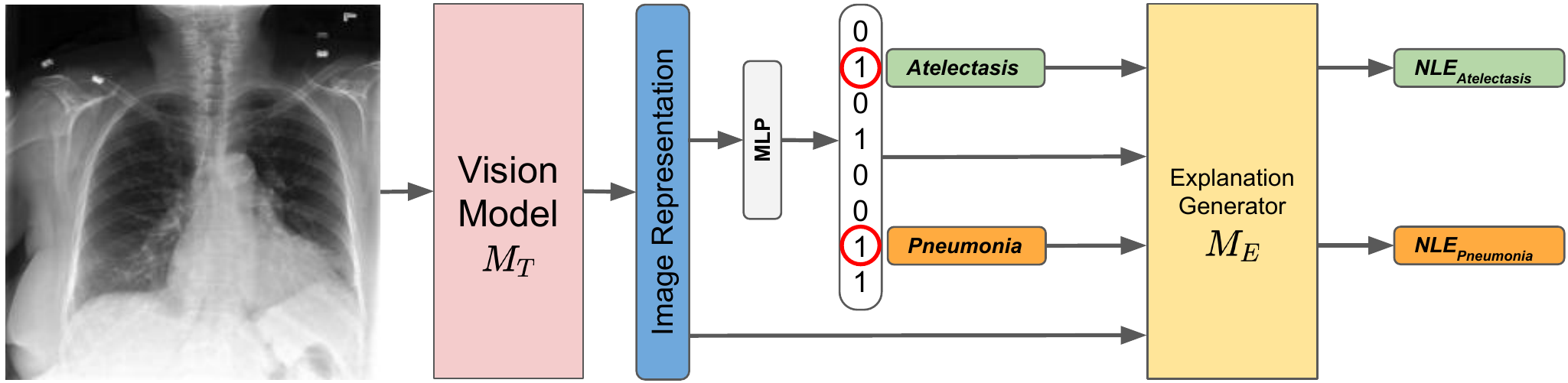}
\end{center}
    \caption{The model pipeline to provide an NLE for a prediction. 
    }%
    \label{fig:nle_approach}
\end{figure}

\section{Experimental Setup}
We evaluate three baselines on our MIMIC-NLE dataset. We propose both automatic and expert-based evaluations.

\smallskip \noindent \textbf{Baselines.} First, we adapt two state-of-the-art chest X-ray captioning architectures to follow the NLE generation approach outlined in Fig.\ \ref{fig:nle_approach}. We re-implement TieNet \cite{wang_tienet_2018}, following their paper and a publicly available re-implementation\footnote{\url{https://github.com/farrell236/RATCHET/tree/tienet}} and RATCHET~\cite{hou_ratchet_2021}, a recent transformer-based captioning method. 
For each approach, we use DenseNet-121 \cite{Huang_2017_CVPR} as the vision model $M_T$. DenseNet-121 is a convolutional neural networks (CNN) widely used for chest X-ray classification \cite{irvin_chexpert_2019}.
For TieNet, contrary to the original model, we do not predict the labels from the learned text representations, but condition the NLE generation on the labels. TieNet's decoder, an LSTM \cite{hochreiter_long_1997}, is conditioned on the pooled features of the image $x_{\text{REP}}$ via an attention layer, on the prediction vector $Y$ by adding its embedding to $x_{\text{REP}}$, and on the diagnosis by initializing the hidden state of the decoder with an encoding of the predicted diagnosis $p_j$ in textual form. RATCHET considers $49 \times 1024$ feature maps as $x_{\text{REP}}$, and the embedding of $Y$ is modeled as an additional, $50$-th, feature map. We condition on $p_j$ in the same way as in TieNet, by initializing the hidden state of the decoder. 

We also propose an additional baseline referred to as DPT (\textbf{D}enseNet-121 + G\textbf{PT}-2), which is inspired by state-of-the-art NLE approaches for natural image understanding  \cite{marasovic_natural_2020, kayser_e-vil_2021}. DPT leverages a DenseNet-121 as $M_T$ and a vision-enabled GPT-2 language model \cite{radford_language} as $M_E$. An attention layer enforces $M_E$ 
to focus on the same regions in the scan as $M_T$. 
We get attention-weighted $7 \times 7 \times 1024$ feature maps from the last layer of DenseNet-121. We consider the NLE generation as a sequence-to-sequence translation from the 49 feature maps (to which we add the relative position of each feature map as a position encoding), a token for $Y$, and the diagnosis $p_j$ in text form. 
For fair comparison, we use the same GPT-2 vocabulary for all models and initialize all word embeddings with the pre-trained GPT-2 weights. All models use DenseNet-121 from TorchXRayVision \cite{Cohen2020xrv}, which was pre-trained on CheXpert \cite{irvin_chexpert_2019}. All text generation is done via greedy decoding. 

\smallskip\noindent \textbf{Model Training.} 
We use a maximum sequence length of 38 during training, which corresponds to the 98th percentile of the training set. We use the Adam optimizer with weight decay for the transformer models and without weight decay for TieNet (as it failed otherwise). We also use a linear scheduler with warmup for the learning rate. For each model, we experiment with different learning rates and batch sizes as hyperparameters. 
Using the dev set, we obtain the best hyperparameters as follows: for DPT, a learning rate of $5 \times 10^{-4}$ and batch size of 16; for RATCHET, a batch size 16 and learning rate $5 \times 10^{-5}$; for TieNet, a batch size of 32 and learning rate of $1 \times 10^{-3}$. We selected these 
based on the product of the task score $S_T$, $\mathrm{CLEV}$ score, and the average of the BERTScore \cite{zhang_bertscore_2019} and METEOR \cite{banerjee_meteor_2005} score, all outlined below.

\smallskip\noindent \textbf{Evaluation Metrics.} We evaluate models both on their ability to solve the task, i.e., image multi-label classification, and their ability to explain how they solved the task, i.e., providing NLEs. The task score $S_T$ is given by the weighted AUC score, where we consider \emph{uncertain} and \emph{positive} as one class (following \citet{irvin_chexpert_2019}). For the NLEs scores, we only consider the NLEs that explain correctly predicted labels, as in \cite{camburu_e-snli_2018, kayser_e-vil_2021}. Previous works have shown that automated natural language generation (NLG) metrics are underperforming for NLEs, as the same answers can be explained in different syntactic forms and even different semantic meanings \cite{kayser_e-vil_2021}. We therefore propose the $\mathrm{CLEV}$ (CLinical EVidence) score, which verifies whether an NLE refers to the right clinical evidence. For this, we leverage the CheXbert labeler, which extracts the evidence labels referred to in the GT and generated NLEs. For example, if the GT NLE mentions \emph{Lung Opacity} and \emph{Consolidation} as evidence for \emph{Pneumonia}, we expect the generated NLE to contain the same findings. The $\mathrm{CLEV}$ score is the accuracy over all the generated NLEs, i.e., what share of them contains exactly the same evidence labels as the corresponding GT. We also provide the NLG metrics of
BLEU, Rouge, CIDEr, SPICE, BERTScore, and METEOR as in \cite{kayser_e-vil_2021}. For BERTScore, we initialize the weights with a clinical text pretrained BERT model.\footnote{\url{https://huggingface.co/emilyalsentzer/Bio\_ClinicalBERT}} 
It is worth noting that, out of the given suite of NLG metrics, BERTScore and METEOR have previously been shown to have the highest (although still low in absolute value) empirical correlation to human judgment in natural image tasks \cite{kayser_e-vil_2021}.

\smallskip\noindent \textbf{Clinical Evaluation.} Given the difficulties of evaluating NLEs with automatic NLG metrics, we also provide an assessment of the NLEs by a clinician. They were presented with 50 X-ray scans, a diagnosis to be explained, and four different NLEs: the GT and one for each of our three models. The NLEs are shuffled for every image, and the clinician does not know which is which. They are then asked to judge how well each NLE explains the diagnosis given the image on a Likert scale of 1 (\emph{very bad}) to 5 (\emph{very good}).

\section{Results and Discussion}

    

\begin{table*}[t!]
    \begin{center}
    \caption{The $S_T$ score, clinical evaluation, and NLG scores for our baselines on the MIMIC-NLE test set. $\geq$GT reflects the share of generated NLEs that received a rating on-par or better than the GT. Clin.Sc.\ reflects the average rating of 1 (lowest) to 5 (highest) that was given to the NLEs by a clinician. R-L refers to Rouge-L, and B$n$ to the $n$-gram BLEU scores. Best results are in bold. As we only evaluate NLEs for correctly predicted diagnoses, our NLG metrics cover 534, 560, and 490 explanations for RATCHET, TieNet, and DPT, respectively.}
    \begin{tabulary}{\textwidth}{lccccccccccc}
    \toprule
                 & AUC  & $\geq$GT  & Clin.Sc. & CLEV & BERTS. & MET. & B1   & B4  & R-L  & CIDEr & SPICE \\ \midrule
    GT           & -    & -                   & 3.20           & -    & -      & -    & -    & -   & -    & -  & -  \\
    DenseNet-121 & 65.2 & -                   & -              & -    & -      & -    & -    & -   & -    & -  & -  \\
    RATCHET      & \textbf{66.4} & \textbf{48\% }               & \textbf{2.90}           & 74.7 & 77.6   & \textbf{14.1} & \textbf{22.5} & \textbf{4.7} & \textbf{22.2} & \textbf{37.9} & \textbf{20.0} \\
    TieNet       & 64.6 & 40\%                & 2.60           & \textbf{78.0} & \textbf{78.0}   & 12.4 & 17.3 & 3.5 & 19.4 & 33.9 & 17.2 \\
    DPT          & 62.5 & \textbf{48\%}                & 2.66           & 74.9 & 77.3   & 11.3 & 17.5 & 2.4 & 15.4 & 17.4 & 13.7 \\
    \bottomrule
    \end{tabulary}
    \end{center}
    \label{res:all}
\end{table*}

Table~\ref{res:all} contains the AUC task score, the evaluation conducted by a clinician, and the automatic NLG metrics. We also provide AUC results for DenseNet-121 only, i.e., without an $M_E$ module, which shows that providing NLEs can improve task performance, e.g., for RATCHET. 
We observed that the main reason why GT NLEs obtain an absolute rating score of 3.2/5 is inter-annotator disagreement between our clinician and the author of the reports (sometimes due to lack of patient context given in this scenario).
Another reason is that some of the GT NLEs refer to a change in pathology 
with respect to a previous study, which is not something that can be assessed from the image. 
We also observe that the CLEV score neither correlates to expert evaluation nor NLG metrics. One explanation could be that the evidence labels in MIMIC-NLE are highly imbalanced, i.e., predominantly \emph{Lung Opacity}. Therefore, as the CLEV score does not take into account much of the diversity that is inherent in our NLEs, such as the location of findings, and their size and appearance, a model that generates generic NLEs that make reference to \emph{Lung Opacity} will yield a good CLEV score, as was the case for TieNet.
Overall, the NLG metrics are on-par with the NLG metrics for report generation \cite{hou_ratchet_2021}. Hence, the difficulty of generating longer texts (reports) seems to be offset by the degree of difficulty of specific, but shorter NLEs. The results also indicate that the NLG metrics are generally poor at reflecting expert judgment. More precisely, BERTScore, which showed the highest correlation with human judgment for natural images \cite{kayser_e-vil_2021}, has poor indicative qualities for medical NLEs. One reason 
could be that automatic NLG metrics generally put equal emphasis on most words, while certain keywords, such as the location of a finding, 
can make an NLE clinically wrong, but only contribute little to the NLG score. While the GPT-2 based architecture of DPT proved very efficient for natural images \cite{kayser_e-vil_2021}, its performance is less convincing on medical images. A reason could be that GPT-2 is too large and relies on embedded commonsense knowledge to generate NLEs, which is less helpful on highly specific medical text. Example NLE generations are provided in the appendix.

\section{Summary and Outlook}
    In this work, we introduced MIMIC-NLE, the first dataset of NLEs in the medical domain. We proposed and validated three baselines on MIMIC-NLE. 
Providing NLEs for medical imaging is a challenging and worthwhile task that is far from being solved, and we hope our contribution paves the way for future work. Open tasks include 
providing robust automatic metrics for NLEs and introducing 
more medical NLE datasets and better performing NLE models. 





\section*{Acknowledgments}
    We thank Sarim Ather for useful discussions and feedback. M.K. is supported by the EPSRC Center for Doctoral Training in Health Data Science (EP/S02428X/1), and by Elsevier BV. This work has been partially funded by the ERC (853489---DEXIM) and by the DFG (2064/1---Project number 390727645). This work has also been supported by the Alan Turing Institute under the EPSRC
grant EP/N510129/1,  by the AXA Research Fund,  the ESRC
grant “Unlocking the Potential of AI for English Law”,  the EPSRC grant EP/R013667/1, 
and by the EU TAILOR grant. We also acknowledge the use of GPU computing support by Scan Computers International Ltd.
BWP acknowledges a Nuffield Department of Population Health Research Fellowship. OMC acknowledges a Leverhulme Early Career Fellowship.

\bibliographystyle{splncs04nat}
\bibliography{My_Library.bib}

\clearpage
\begin{appendix}
    \section{Supplementary} \label{app:data}
        \begin{table}[]
\caption{Our different filters that were applied to extract image-diagnosis-NLE triplets from MIMIC-CXR. \# of sentences corresponds to unique image-sentence pairs. ``Non-descriptive aspects'' are aspects that cannot be derived from the image itself. Duplicates are sentences that are from the same report and mention the same labels.}
\label{tab:my-table}
\resizebox{\textwidth}{!}{%
\begin{tabular}{|l|r|}
\hline
Filter                                                           & \multicolumn{1}{l|}{\# of sentences} \\ \hline
Extract Findings and Impression sections                         & 1,383,533                            \\
Remove sentences with anonymized data                            & 1,304,465                            \\
\begin{tabular}[c]{@{}l@{}}Remove sentences referring to non-descriptive aspects: \\ \ \textbf{Patient history}: ``prior'', ``compare'', ``change'', ``deteriorat'', ``increase'', \\ \ ``decrease'', ``previous'', ``patient''\\ \ \textbf{Recommendations}: ``recommend'', ``perform'', ``follow''\\ \ \textbf{Technical}: ``CT'', ``technique'', `` position'', ``exam'', ``assess'', ``view'', ``imag''\end{tabular} &
  1,007,002 \\
Filter by rules from Table \ref{tab:lbl_rules} & 43,612                               \\
Remove duplicates                                                & 39,094                               \\
Remove studies without AP or PA images                           & 38,003                               \\ \hline
\end{tabular}%
}
\end{table}

\begin{table}[]
\centering
\caption{The distribution of diagnosis labels in all our NLEs. Diagnosis label combinations are ordered by occurrence. \# of sentences corresponds to unique image-sentence pairs. The table is displayed in two halves.}
\label{tab:diaglbl}
\resizebox{\textwidth}{!}{%
\begin{tabular}{|l|r||l|r|}
\hline
Diagnosis Labels                & \# of sentences & Diagnosis Labels               & \# of sentences \\
\hline
Atelectasis                     & 10,616          & Atel., Edema, Pneumonia        & 104             \\
Pneumonia                       & 9,032           & Atelectasis, Edema             & 103             \\
Edema                           & 5,098           & Pl. Eff., Pneumothorax         & 65              \\
Atelectasis, Pneumonia          & 4,773           & Pleural Effusion, Pneumonia    & 55              \\
Pleural Effusion                & 4,585           & Edema, Pleural Effusion        & 31              \\
Consolidation                   & 916             & Atelectasis, Pleural Other     & 17              \\
Pleural Other                   & 846             & Atel., Edema, Pl. Eff.         & 9               \\
Edema, Pneumonia                & 623             & Edema, Pl. Eff., Pneumonia     & 8               \\
Atelectasis, Pleural Effusion   & 397             & Atel., Edema, Pl. Eff., Pneum. & 6               \\
Pneumothorax                    & 311             & Edema, Pleural Other           & 6               \\
Pleural Effusion, Pleural Other & 195             & Pleural Other, Pneumonia       & 4               \\
Atel., Pl. Eff., Pneumonia      & 190             & Other                          & 13              \\ 
\hline
\end{tabular}%
}
\end{table}

\begin{table}[]
\centering
\caption{The distribution of evidence labels in all our NLEs. Evidence label combinations are ordered by occurrence. \# of sentences corresponds to unique image-sentence pairs. The table is displayed in two halves.}
\label{tab:evlbl}
\begin{tabular}{|l|r||l|r|}
\hline
Evidence Labels & \# of sentences & Evidence Labels             & \# of sentences \\
\hline
Lung Opacity    & 29,115          & Consolidation, Lung Opacity & 692             \\
Other / misc.   & 5,842           & Enlarged Cardiomediastinum  & 141             \\
Consolidation   & 2,102           & Lung Lesion                 & 111             \\ \hline
\end{tabular}
\end{table}

\begin{table}[]
\centering
\caption{The occurrence of different explanation keywords in the unfiltered sentences from MIMIC-CXR. The table is displayed in two halves.}
\label{tab:explkw}
\begin{tabular}{|l|r||l|r|}
\hline
Keyword          & Count                     & Keyword        & Count \\
\hline
suggest          & 19,787                     & relate         & 5,241                      \\
reflect          & 17,878                     & may represent  & 5,141                      \\
due              & 17,771                     & potentially    & 3,160                      \\
consistent with  & 14,654                     & worrisome for  & 1,685                      \\
concerning for   & 10,243                     & indicate       & 1,568                      \\
compatible with  & 6,025                      & account        & 1,193                      \\
likely represent & 5,239                      & suspicious for & 684                       \\
\hline
\end{tabular}
\end{table}

\begin{figure}
\begin{center}
\includegraphics[width=1\linewidth]{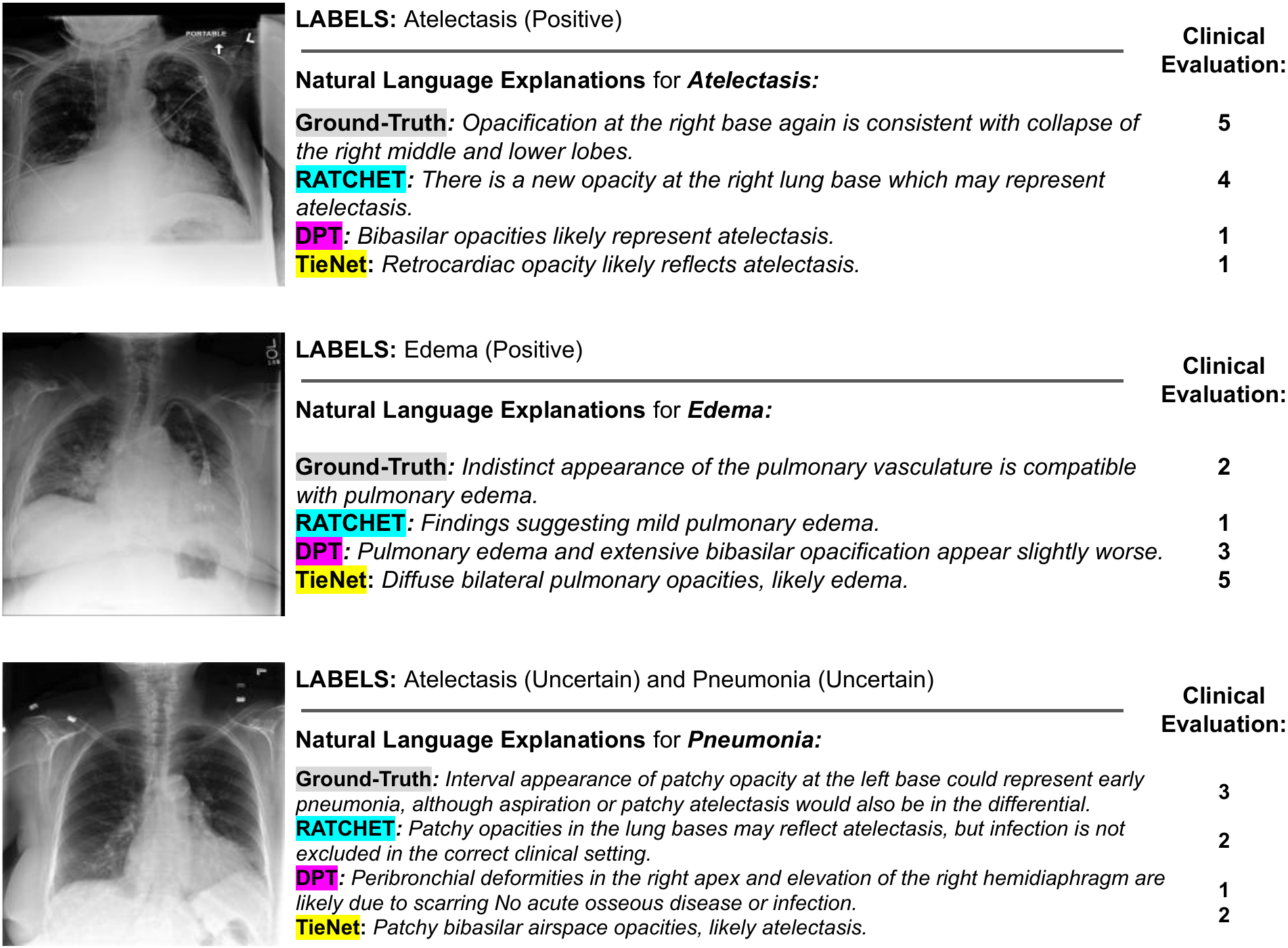}
\end{center}
    \caption{The GT NLEs and three model-generated NLEs explaining three different diagnosis labels on three different scans. The clinical evaluation is given on a Likert scale, where 5 is the highest, and 1 is the lowest score.}%
    \label{fig:nle_example}
\end{figure}
\end{appendix}

\end{document}